# How Artificial Intelligence LLM Engines Shape the Global Conflict Information Environment

Jason Miklian
Research Professor
Centre for Global Sustainability, University of Oslo
Jason.miklian@globe.uio.no

**Abstract**

Artificial Intelligence (AI) answer engines now field a growing share of the questions that analysts, scholars, and the public ask about issues of peace and conflict. Large Language Models (LLMs) are known to hallucinate under certain conditions, but do these errors have discernible patterns when they are asked about conflicts, and if so what can that teach us about the changing global conflict information environment? To answer, we first asked a battery of questions about 28 conflicts to five leading answer engines and scored their 5,460 answers against documented evidence. We found that the thinner the retrievable record around a given conflict, the more the engines invent, misattribute, and miscount. Thin records don't just encourage hallucination, but create structural exposure to mis- and disinformation, because they are the easiest records to warp through Generative Engine Optimization (GEO) to bias engine responses. Through an analysis of 1,047 websites that the AI LLMs pulled conflict facts from, we found that GEO source optimization is already happening, and while state-partisan digital capture remains incipient it is rapidly growing. We explain what these findings mean for scholarship with the rise of GEO information warfare, and for policy argue for a return to the deep local monitoring and translation-based research that AI tools cannot replicate, closing with a discussion of future research opportunities and challenges in this fast-moving space.

## Introduction

Ask any of the major Artificial Intelligence Large Language Model (AI LLM) answer engines how many civilians were killed in Ukraine in 2025 and they converge around the same documented figure. OpenAI, Grok, Perplexity, and Anthropic return the UN monitoring mission's count of at least 2,514, while Gemini returns at least 2,526, against a documented range for the year of 2,514 to 3,000. The answer is typically sourced to the UN or other authoritative sources, hedged for underreporting, and generally correct to a Wikipedia-level of informational accuracy. But ask the same engines about Mobondo (Teke-Yaka) intercommunal violence in western Democratic Republic of the Congo, and the engines guess at figures, refuse to give an answer, make up numbers, or respond by talking about entirely different conflicts, leaning upon the closest matches in their training data to generate plausible answers that have little connection to reality.

A growing share of scholars, professionals and the public who want quick facts about a conflict now ask a chatbot rather than conducting deeper research, which retrieves what has been written and compresses it into a sentence or two. Independent testing gives reason to distrust the compression: LLMs cite the wrong source for summarized news more than sixty percent of the time (Jaźwińska and Chandrasekar 2025), usually because of challenges in easily finding an authoritative source (Suzgun et al. 2026). Therefore, does growing reliance on AI LLMs risk that we will become even less informed on low-information conflicts where media attention is sparse? Organized violence killed more people in 2025 than in any year since 1994 except 2022 (Davies et al. 2026), and much of that violence drew little notice. Combined with a global retrenchment in media diversity and creation, if there is increasingly less information to draw upon and less time for individuals to research themselves, what might this mean for our understanding of conflicts?

To answer, we asked five leading AI LLMs about 28 different conflicts to assess their responses and the material they drew from. This work delivered three findings. First, the engines are least accurate on the least-covered conflicts: the thinner the record around a conflict, the more often they invent, misattribute, and miscount, and so-called "forgotten conflicts" come off worst. Second, the engines are missing settled, knowable facts precisely where coverage is thin, a failure to find what is known rather than simply asking unanswerable questions. Third, the thinnest records are the most engineered to be retrieved by AI LLMs through Generative Engine Optimization tools. While much of that engineering comes from small local outlets chasing visibility and self-promotion, propaganda operations are beginning to game AI engines to shape what people see when asked about the "truth" of low-information conflicts. Put together, we find that all leading AI LLMs amplify the world's neglect on most conflicts, and conclude with a discussion of what would repair these failures.

## Amplification, Not Just Hallucination: A Literature Review

These topics interconnect through a series of literatures. The first has established that large language models invent facts. Hallucination is now catalogued as a systematic failure mode rather than an occasional glitch (Ji et al. 2023), benchmarked against the human falsehoods models are prone to imitate (Lin et al. 2022), and traced in part to the training objective itself, which rewards fluent plausibility over grounded accuracy (Bender et al. 2021). Detection methods can estimate when a model is confabulating (Farquhar et al. 2024), and taxonomies of model risk place factual error alongside manipulation and bias as first-order harms (Weidinger et al. 2022; Augenstein et al. 2024). Moreover, fabricated citations show that models invent not only facts but the very sources that would anchor them (Walters and Wilder 2023), and scale does not reliably rescue accuracy: larger, more instructable models can grow *less* honest about the limits of their knowledge, answering confidently where a smaller model would have balked (Zhou et al. 2024).

The same isolation marks parallel work on ideological slant, which audits a model's own leanings on abstract political prompts (Motoki et al. 2024; Rozado 2024) rather than asking how a skewed record bends a grounded answer. When engines cite their sources, they often cite ones that do not support the claim (Liu et al. 2023; Jaźwińska and Chandrasekar 2025); most errors fall under issues of retrieval rather than reasoning since an engine that reaches the right source usually reads it correctly (Suzgun et al. 2026). The closest antecedent to our study puts three LLM chatbots against disinformation narratives about a single heavily documented war, Russia's invasion of Ukraine, and finds accuracy uneven across languages and unstable over time (Makhortykh et al. 2024). Of course, most of this work tests models in isolation, on general knowledge topics, in English. What happens when LLMs tackle issues with the thinness of records themselves, on issues of deep importance but low information? Understanding this issue in a comparative manner is a key knowledge gap.

A second body of literature explains why a thin record can be actively dangerous rather than merely incomplete. The concept of the data void typifies a forgotten conflict: a query for which few authoritative sources exist, so that whatever thin or self-interested material is present fills the space and is easily exploited and exploitable (Golebiewski and boyd 2019). Work on information disorder shows how readily false content outruns true content online (Vosoughi et al. 2018; Lazer et al. 2018), how automation amplifies low-credibility sources (Shao et al. 2018), and how the very act of searching to verify a claim can raise its perceived truth when the returned evidence is thin (Aslett et al. 2024). Generative models lower the cost of producing persuasive text at scale for influence operations (Kreps et al. 2022; Goldstein et al. 2023) and invite adversarial optimization aimed squarely at what the engines will repeat (Nestaas et al. 2024; Aggarwal et al. 2024). That optimization is already happening: pro-Kremlin networks flood the open web with millions of articles aimed at contaminating what models train on and retrieve, grooming the machines rather than persuading readers (Padalko 2025).

This literature anticipates the rise of vulnerability on low-information conflicts but has not connected the data void to empirical accounts of *which* topics are most exposed, nor asked systematically who is doing the optimizing on the ground. The thinnest conflict media records are also the most concentrated locally (including mass market propaganda run by conflict actors themselves) and more likely to be quoted by AI as fact in the absence of authoritative third party sources. Moreover, Generative Engine Optimization (GEO) has arisen in digital marketing, an advancement of Search Engine Optimization (SEO) (to promote links on sites like Google) to try to game LLMs by putting their material at the forefront of responses to queries. Two knowledge gaps arise: are LLMs pulling biased or imbalanced material on conflicts, and if so is this the result of actors attempting to game the response system?

A third literature explains why the record that LLMs pull from is thin to begin with. It is a frustrating but durable cliche that global news does not track suffering but the criteria of newsworthiness, so violence far from centres of power and interest goes uncovered almost regardless of scale (Galtung and Ruge 1965; Hawkins 2016). News attention concentrates on the rare, dramatic incident rather than the underlying distribution of violence, so what audiences see systematically misrepresents where harm actually falls, as coverage of cyber conflict shows (Makridis et al. 2024). The condition has worsened as the economics of journalism have collapsed, hollowing out local and international reporting alike and leaving local journalism in market failure (Usher 2023), eroding public trust in and engagement with news (Strömbäck et al. 2020; Kozyreva et al. 2020). Conflict measurement describes the upstream consequence: violence is undercounted precisely where coverage is sparse (Weidmann 2016), state killings are folded into ordinary combat (Broache et al. 2025), and the quality of a record must be judged before estimates can be trusted (Gohdes and Price 2013). Fatality counts depend on who is present to observe the killing, so where coverage is thinnest the missing dead leave the casualty record radically incomplete (Dawkins 2021).

Uniting these literatures, we see how conflict actors exploit these tensions and voids. Visibility has never been a neutral by-product of publishing, and states learned this early. Junk and partisan outlets game search rankings deliberately, and SEO puts them in front of readers (Bradshaw 2019). Armed groups' own social media output is purposive messaging aimed at mobilizing audiences, and it increasingly constitutes the data that scholars use (Bestvater and Loyle 2025). Some states manufacture online content at industrial scale (hundreds of millions of posts a year) to flood the digital zone and steer the information environment (King et al. 2017), a practice known as "LLM Grooming" in the AI era. Combatants themselves invest strategically in online propaganda, so the digital record of a civil war is often biased (Walter and Phillips 2025).

A state can also control the retrieval infrastructure itself. For example, Russia's state-controlled Yandex tilted the sources readers saw during an anti-regime protest toward state-aligned accounts (Kravets and Toepfl 2022), and search results on contested Soviet-era mass violence align with identifiable sides of the underlying memory war (Makhortykh et al. 2022). Retrieval systems can be captured from above by owning the engine, but also gamed from below by generating content for web crawlers. The models themselves are also contested ground: frontier systems carry geopolitical leanings that track their home states, e.g. DeepSeek's preferences for Chinese state perspectives compared to GPT-4o's softer Western framings (Pacheco et al. 2026), and this favoritism can deliver positively framed political misinformation (Chang et al. 2025). But does the shift from SEO (for search engines) to GEO (for AI LLMs) change whether the state and partisan outlets that mastered the old game have re-tooled for the new information conditions amplified by the rise of AI?

AI answer engines inherit the documentation gap that the political economy of news produces, favor the most retrievable sources over the most authoritative ones, and compress the result into a confident paragraph delivered as settled fact. Reading the reliability, manipulation, and forgotten-conflict literatures together yields four questions. First, do answer engines amplify misinformation about, and inattention to, armed conflicts? Second, is the problem measurable, and how large is it when answers are scored against documented evidence? Third, does reliability depend on how closely the world watches a conflict, tracking the information environment rather than the technology? Fourth, does failure vary across engines or recur within all of them, distinguishing the faults of particular products from properties of retrieval-grounded generation itself?

## Method and data

This study built a thirteen-question battery to ask five leading answer engines about each of twenty-eight different conflicts, repeating every question three times, for 5,460 answers in total in the 15–20 June 2026 period. Each engine ran with web search or grounding switched on, so the answers reflect the retrieve-and-summarize product a typical user would actually see, recording the model version, the time, the full answer, and the sources cited. The run used gpt-5-mini, claude-sonnet-4-6, gemini-3.5-flash, sonar-pro, and grok-4.3, with robustness tests of higher value models (e.g. Claude Opus 4.8, GPT 5.5). Google retired gemini-3-pro during data collection, so the Gemini results reported here run on its successor, gemini-3.5-flash. The thirteen questions (see Appendix) covered a variety of conflict-relevant questions like asking the number of casualties, who did what / who was responsible, displacement actions, and the like. Several were deliberately leading, pushing a high or a low premise, so we could see whether a thin record makes an engine more willing to swallow a false assumption.

The twenty-eight conflicts (see Appendix) span a wide range of attention, from wars that lead newscasts to ones that reach almost no English-language reader. The three attention tiers were assigned at pre-registration on the volume of English-language news attention each conflict attracts, and were frozen in the deposited design: five watched, eight middle, fifteen forgotten. Five are high-attention, among them Ukraine, Gaza, and Sudan; eight are mid-informational; and fifteen can be considered “forgotten”, from Cabo Delgado and Tigray to West Papua, Western Sahara, and Nigeria’s Middle Belt. These labels are coarse, because a conflict can be loud and badly sourced or quiet and decently sourced (Galtung and Ruge 1965), so the main analysis leans on a measured score for how thinly each conflict is covered rather than on the labels.

That score captures the information environment around each conflict, and it combines two things an engine can draw on. The first is the online news pool, taken from GDELT’s Global Knowledge Graph, of how many articles exist, across how many distinct websites, and how concentrated they are. The second is on-the-ground reporting, taken from ACLED’s record of roughly half a million events from January 2023 to 15 June 2025 in each conflict’s region: how many distinct source combinations report the violence, and how concentrated those sources are (Raleigh et al. 2010). Each country’s press conditions, from V-Dem, sit alongside these as the measures the within-country comparisons below hold fixed. The combined score runs from saturated coverage, as in Mexico, to near-absence, as in western Congo, with a higher score meaning a thinner record. Four conflict pairs sit in a single country to enable more direct comparison. The Democratic Republic of the Congo gives the sharpest pair, the closely watched M23 war in the east against the Mobondo (Teke-Yaka) intercommunal violence in the west. India pairs Kashmir against Manipur, Pakistan pairs Kashmir against Balochistan, and Nigeria pairs Lake Chad against the Middle Belt. The Pakistan pair holds Pakistani press freedom fixed only nominally, because Kashmir's information-environment measures are coded on the Indian side.

For 133 question-conflict keys we built a documented answer range or status from recorded sources (UCDP, the UN bodies OHCHR and OCHA, IOM, and UNHCR). Some carry a defensible single value, such as a displacement total, and some are contested but bounded, a credible range with no agreed point, which is where most death tolls sit. Others are disputed, such as which side committed a massacre or whether a killing campaign crosses the genocide threshold; for these the right answer is to lay out the dispute, and an engine that picks a side and states it as fact is marked wrong even when its number is plausible. A last group has no settled figure at all because the underlying conflict record is itself biased: it undercounts violence where coverage is thin (Weidmann 2016) and tends to log civilian killings by state forces as ordinary combat rather than deliberate targeting (Broache et al. 2025). Scoring an engine against a single hard number would import those biases, so we score against an evidenced range (Gohdes and Price 2013). UCDP anchors the ranges because its mapped event data hold up better for local scoring than ACLED, which has uneven quality control below the national level (Eck 2012; Sundberg and Melander 2013). The sources follow a priority order fixed in the pre-registration: UCDP for battle-related totals, OHCHR for verified civilian counts, UN fact-finding mechanisms for attribution, and IOM, OCHA, and UNHCR for displacement and siege. Attribution counts as settled only where a UN-level inquiry has fixed responsibility, which is why Gaza and Sudan carry settled attributions while Kashmir remains contested. That rule is an epistemic choice, and it means the benchmark inherits the geography of international monitoring, which itself tracks attention: the same instruments that verify Ukraine's dead to the individual are weaker in forgotten wars.

Each answer is scored on a five-point scale: a figure inside the documented range, stated with appropriate caution, scores +2; a figure close to the range but asserted as precise scores +1; the right order of magnitude outside the range scores 0; an answer wrong by an order of magnitude, or a disputed attribution stated as settled, scores −1; and an invented figure or named culprit that no source supports scores −2. Throughout, an answer counts as an error when it scores zero or below, meaning at best it is only roughly right. On facts with no settled figure we also record separately whether the engine says so or asserts a number anyway.

Two scoring passes assessed the 5,460 answers. The first was a deterministic rule scorer that handled bounded numeric facts: it pulled the engine's stated figure out of the text and compares it to the frozen interval. The second is a single automated pass that applied the whole codebook to each answer. That coder was withheld the engine's identity and the conflict's attention tier, but a conflict's name is often visible inside the answer text, so its blindness is partial rather than complete. As a reliability check on the numeric facts, on 2,386 bounded numeric answers the scorers agreed 44.4 percent of the time and within one point 78.0 percent, correlating at 0.448; and the conflict-blind rule scorer, which carries no codebook priors, reproduces the headline thinness gradient on its own and more steeply (−0.369, conflict-clustered p=0.007), which bounds the concern that coder priors drive the numeric result. Two reference comparisons sit outside the main results: Anthropic's questions rerun on its stronger Opus model, and a spot-check of OpenAI's flagship against the smaller model used in the main run.

Across the tables that follow, a higher thinness score means a thinner record; accuracy is the mean of the five-point score, where +2 is best and −2 worst; and an answer counts as an error when it scores zero or below. P-values come from models with engine effects and standard errors grouped by conflict unless noted, and “significant” is used in its statistical sense. Because the predictor varies across only twenty-eight conflicts, wild cluster bootstrap p-values (Rademacher weights, null imposed, 9,999 replications) are reported for the headline models and serve as the reference inference, with the conflict-level regression as the primary evidence for the gradient. See Appendix for full data processing and replication information.

## Results

We have three main findings. First and perhaps most expected, AI is least accurate on the conflicts the world does not watch. The error rate shows it clearly: the engines miss 21.0% of the time on Ukraine but miss 42.6% on South Sudan, 48.7% on Lake Chad, and 43.6% on western Congo. The decline is gradual from most- to least- covered conflicts, from about +1.0 on the five-point scale for the best-covered third of conflicts to about +0.7 for the thinnest third. Allowing for the fact that some engines may simply be better than others, each step toward a thinner record costs a consistent amount of accuracy, c. 0.18 points per step. The gradient holds at the level of the conflicts (28 conflicts, $p=0.024$; Spearman rank correlation −0.41, $p=0.033$), while at the level of individual answers it is borderline after accounting for the small number of conflict clusters (wild cluster bootstrap $p=0.055$). The predicted swing from the thickest record to the thinnest is about half a point on the five-point scale (R2 of 0.02 at the answer level and 0.21 at the conflict level), and sorting the conflicts into watched, middling, and forgotten tiers instead gives the same slide, from about +1.1 for the watched conflicts to about +0.8 for the forgotten (see Table 1).

The obvious objection is geography: poor countries have poorer reporting, so this could be a story about regions. So we restricted the comparison to conflicts within the same world region, which did not change the results. Another concern is censorship, that thinly covered conflicts simply sit in more repressive states. Here we take the Democratic Republic of the Congo as an illustration. The war in the east around Goma is followed closely; the Mobondo (Teke-Yaka) intercommunal killings in the west, in the same country with the same press laws in the same year, are not. The question prompts used the "Mai-Mai" label common in earlier coverage of the area; engines that corrected the terminology to the Mobondo (Teke-Yaka) conflict in their answers show the questions remained answerable, and the naming is corrected throughout this article. The east draws around nine thousand news articles a year from nearly nine hundred distinct outlets, the west about seven hundred articles from 136. Across all five engines the answers about the eastern conflict beat the answers about the western one by c. 0.7 (p=0.016). What the pair shows descriptively, with national press freedom held constant, is how the engines behave where the record is weak: in the forgotten west they assert a single value as settled five times as often (33 percent against 6 percent) and account for every confidently wrong answer in the pair (6 percent against none).

Behind these comparisons sits one consistent relationship: each move toward a thinner record lowers accuracy. The negative sign holds under every check in Table 1, and a larger number means a steeper drop. For the explicitly dated 2025 casualty questions, the slope is +0.096 (p=0.225; N=840). Regarding the other three within-country pairs, Kashmir is answered slightly worse than Manipur (gap −0.297, p=0.18) and worse than Balochistan (−0.287, p=0.14), and Lake Chad slightly worse than the Middle Belt (−0.149, p=0.30). In each, the more-watched theatre scores lower, and none of the three reaches significance. The two Kashmir reversals reflect the dense but contested record taken up in the Kashmir passage below, while the Nigeria pair sets two forgotten theatres against each other.

**Table 1. Accuracy changes as a function of coverage**

| Measure | Detail | Value |
|---|---|---|
| **A. Mean accuracy by coverage and attention  (+2 best to −2 worst)** | | |
| Best-covered third (by thinness) | | +1.02 |
| Middle third | | +0.92 |
| Thinnest third | | +0.71 |
| Watched tier | N=975; error rate 28.1% | +1.11 |
| Middle tier | N=1,560; error rate 31.5% | +0.93 |
| Forgotten tier | N=2,925; error rate 36.9% | +0.79 |
| **B. The gradient under different models  (estimate on thinness; 28 conflict clusters)** | | |
| Accuracy on thinness | engine effects, grouped by conflict | −0.175 (0.072), p=0.015; bootstrap p=0.055 |
| Probability of error (score ≤ 0) on thinness | | +0.054 (0.020), p=0.007; bootstrap p=0.035 |

| Measure | Detail | Value |
|---|---|---|
| Accuracy on thinness, conflict level | N=28 conflict means; HC1 | −0.175 (0.073), p=0.024; Spearman −0.405, p=0.033 |
| Ordered logit, five-category score | cluster bootstrap, 999 reps | −0.270, 95% CI [−0.49, −0.03] |
| Accuracy on thinness | adding world-region effects | −0.441 (0.042), p<0.001 |
| **C. Validity checks: the two scorers, and run-to-run consistency** | | |
| Answers where both scorers overlap | | 2,386 |
| Exact agreement between scorers | | 44.4% |
| Agreement within one point | | 78.0% |
| Correlation between the two scorers | | 0.448 |
| Thinness slope, automatic scorer alone | numeric extraction; conflict-clustered inference | −0.369, p=0.007 |
| Question-cells with at least two repeats | | 1,820 |
| Mean accuracy spread within a cell | | 0.536 |
| Spread on thinness | regression; correlation +0.20 | +0.035, p=0.48 |
| Median run-to-run variation in the numeric figure | | 0.455 |

*Notes. Accuracy is the mean of the five-point score (+2 best, −2 worst); an error is any answer scoring zero or below. Estimates in section B come from models with engine effects and standard errors grouped by conflict unless noted; a negative coefficient means accuracy falls as the record thins. Section C: the automatic scorer extracts the numeric figure and scores it against the frozen range with no knowledge of which conflict it reads; its steeper slope shows the gradient is not an artifact of the codebook scorer. Consistency: spread is the within-cell standard deviation of accuracy across three repeats; the flat slope on thinness shows thin coverage costs accuracy without costing self-consistency. With twenty-eight clusters the analytic cluster-robust p-values are anti-conservative; the wild cluster bootstrap p-values shown alongside them (Rademacher, null imposed, 9,999 replications) are the reference inference, and the conflict-level regression is the primary evidence for the gradient.*

We also check if our grading produced the effect. The automatic scorer read only the number in an answer and knows nothing about which conflict it is scoring. It found the same decline, slightly steeper, so the slide comes from the engines rather than from our codebook. The two scorers agree closely, and the automatic scorer’s own slope is the steeper of the two (Table 1, section C). Regarding run-to-run consistency, we asked each engine the same question three times to determine the spread between its answers. We found it was typically about half a point on the five-point scale, and stays flat as a conflict's record thins (p=0.48). Therefore, thin coverage costs accuracy while leaving the engines' consistency untouched. (Table 1, section C.) This survives our case and measurement checks, staying negative and significant when Mexico, Tigray, Haiti, or Western Sahara is dropped in turn, or all three contested cases at once (p at or below 0.026 throughout, and strengthening without Mexico). It also holds under a principal-components version of the thinness index, under GDELT-only and ACLED-only versions estimated separately, and under every leave-one-component-out variant, with effect sizes between −0.09 and −0.13 per standard deviation against the published −0.125; and it strengthens when scoring is restricted to benchmark-bearing facts (−0.259, p=0.001). The single marginal variant drops the media-concentration term (p=0.060) with the magnitude unchanged. The per-conflict picture, sorted from best-covered to thinnest, shows the same slide. Our findings follow, and Appendix Table F2 reports the full battery:

**Table 1A. Per-conflict results**

| Conflict | Attention | Region | Thinness | Mean accuracy | Error rate | Answerable share | N |
|---|---|---|---|---|---|---|---|
| Mexico cartel violence | Middle | N. America | −1.24 | +0.54 | 42.1% | 0.31 | 195 |
| Sudan (RSF-SAF) | Watched | NE. Africa | −1.05 | +1.39 | 17.9% | 1.00 | 195 |
| Ukraine (Russia-Ukraine) | Watched | E. Europe | −0.95 | +1.33 | 21.0% | 0.92 | 195 |
| Gaza (Israel-Hamas) | Watched | M. East | −0.95 | +1.04 | 30.3% | 0.92 | 195 |
| Yemen | Middle | M. East | −0.80 | +0.59 | 40.5% | 0.31 | 195 |
| Myanmar civil conflict | Middle | SE. Asia | −0.77 | +1.25 | 21.5% | 0.62 | 195 |
| Haiti gang conflict | Middle | Caribbean | −0.75 | +0.97 | 33.8% | 1.00 | 195 |
| Somalia (al-Shabaab) | Middle | E. Africa | −0.68 | +0.94 | 26.7% | 0.92 | 195 |
| Burkina Faso (JNIM) | Middle | W. Africa | −0.22 | +1.08 | 27.2% | 0.69 | 195 |
| Mali (JNIM) | Middle | W. Africa | −0.20 | +1.03 | 31.3% | 1.00 | 195 |
| Balochistan (Pakistan) | Forgotten | S. Asia | −0.17 | +0.85 | 34.9% | 0.62 | 195 |
| Eastern DRC (M23) | Watched | C. Africa | −0.13 | +1.21 | 24.1% | 1.00 | 195 |
| Niger (Sahel) | Middle | W. Africa | −0.05 | +1.06 | 29.2% | 0.92 | 195 |
| Kashmir (India-Pakistan) | Watched | S. Asia | +0.06 | +0.56 | 47.2% | 0.38 | 195 |
| Cabo Delgado (Mozambique) | Forgotten | SE. Africa | +0.07 | +1.10 | 29.7% | 0.54 | 195 |
| Nigeria Middle Belt | Forgotten | W. Africa | +0.14 | +0.77 | 38.5% | 0.92 | 195 |
| Manipur (India) | Forgotten | S. Asia | +0.15 | +0.86 | 33.3% | 0.23 | 195 |

| Conflict | Attention | Region | Thinness | Mean accuracy | Error rate | Answerable share | N |
|---|---|---|---|---|---|---|---|
| Lake Chad / ISWAP | Forgotten | W. Africa | +0.24 | +0.62 | 48.7% | 0.92 | 195 |
| Catatumbo (Colombia) | Forgotten | S. America | +0.36 | +1.25 | 28.7% | 1.00 | 195 |
| Mindanao / NPA (Philippines) | Forgotten | SE. Asia | +0.39 | +0.86 | 30.3% | 0.62 | 195 |
| Tigray (Ethiopia) | Forgotten | NE. Africa | +0.53 | +0.43 | 43.1% | 0.38 | 195 |
| West Papua (Indonesia) | Forgotten | Oceania | +0.54 | +1.00 | 31.8% | 0.92 | 195 |
| Western Sahara | Forgotten | N. Africa | +0.59 | +0.69 | 37.4% | 0.38 | 195 |
| Central African Republic | Forgotten | C. Africa | +0.59 | +1.10 | 25.1% | 0.85 | 195 |
| South Sudan | Forgotten | E. Africa | +0.64 | +0.63 | 42.6% | 0.69 | 195 |
| Southern Thailand | Forgotten | SE. Asia | +1.01 | +0.55 | 45.1% | 0.38 | 195 |
| Ambazonia (Cameroon) | Forgotten | C. Africa | +1.38 | +0.66 | 40.0% | 0.92 | 195 |
| Western DRC (Mobondo) | Forgotten | C. Africa | +1.57 | +0.51 | 43.6% | 0.00 | 195 |

The relationship carries exceptions. For example, Mexico carries the densest record in the set and still scores among the worst, because its drug-conflict killings rarely come with a settled civilian count to find and only a third of its questions have a documented answer at all. Catatumbo and the Central African Republic sit at the opposite corner, thin records the engines handle well.

A comforting explanation would be that neglected conflicts are simply harder, that they raise more questions with no settled answer, so the engines fail where anyone would. To test, we next sorted every question by documentation, finding that the loss of accuracy in thinly covered conflicts falls on facts with a settled or well-bounded figure. Accuracy drops sharply as coverage thins, while on facts with no agreed figure at all coverage makes no measurable difference. Adding a statistical control for how many of a conflict’s questions are answerable delivers similar results.

**Table 1B. Accuracy on thinness**

| Model | Estimate | Std. error | p | N |
|---|---|---|---|---|
| Correlation between thinness and the share of answerable facts | −0.268 | n/a | n/a | 28 conflicts |
| Accuracy on thinness, no control | −0.175 | 0.072 | 0.015 | 5,460 |
| Accuracy on thinness, controlling for answerability | −0.155 | 0.066 | 0.019 | 5,460 |
| The answerability control itself | +0.179 | 0.126 | 0.15 | 5,460 |
| Thinness slope within answerable facts (settled or bounded) | −0.259 | 0.077 | 0.001 | 3,780 |
| Thinness slope within Tier 3 or untiered facts | +0.000 | 0.096 | 1.00 | 1,680 |

This leads to our second main finding: all answer engines we tested get easy, documented facts wrong, and get them wrong precisely where the surrounding coverage is thin. The answers are available and could be retrieved by any competent researcher, but the engines we tested could not find them because the record around them is too sparse to retrieve from. Independent testing of these systems on breaking news reaches the same conclusion: most errors come from failing to land on the right source rather than from reasoning badly once a source is in hand (Suzgun et al. 2026).

The failure is easy to miss because the engines almost never hesitate. Just two of the 5,460 answers refused a question outright, and only about one in twenty flagged that figures were unverified / contested. Moreover, when faced with a question with no clear answer an engine almost always produces a number or names a culprit anyway. About one answer in five is flatly wrong rather than merely imprecise, and roughly one in 160 is an outright fabrication:

## Table 2. Answer variations

| Measure | Share |
|---|---|
| **A. Accuracy score distribution (all 5,460 answers)** | |
| +2 (inside the range, well qualified) | 43.7% |
| +1 (close, but over-precise) | 22.5% |
| 0 (right order of magnitude, outside range) | 13.3% |

| Measure | Share |
|---|---|
| −1 (wrong by an order of magnitude, or disputed stated as settled) | 19.8% |
| −2 (fabricated figure or culprit) | 0.6% |
| **B. Answer form** | |
| Direct answer | 54.7% |
| Hedged around a figure | 39.9% |
| Says it cannot verify | 5.0% |
| Off-topic | 0.4% |
| Refusal | 0.04% (2 of 5,460) |
| **C. Calibration** | |
| Appropriate hedge | 53.6% |
| Surfaces the dispute | 21.7% |
| States a disputed or bounded fact as a settled single value | 18.5% |
| Confidently wrong | 6.1% |
| **D. Uncertainty proxy on Tier 3 or untiered facts (1,680 answers)** | |
| Overall | 69.2% |
| Watched conflicts | 79.3% |
| Middle | 66.7% |
| Forgotten conflicts | 68.8% |
| OpenAI | 78.9% |
| Perplexity | 72.0% |
| Grok | 69.9% |
| Gemini | 66.1% |
| Anthropic | 58.9% |

When a fact truly has no agreed number, the engines state that about two-thirds of the time. They manage it less often for forgotten conflicts than for prolific ones, and they vary widely among themselves, from Anthropic's engine, which conveys the absence about three-fifths of the time, to OpenAI's, which does so about four-fifths. We score this generously, counting a careful hedge as good enough, so even two-thirds flatters the engines. The failures run from the stale to the surreal.

Asked which group had killed the most civilians in Catatumbo, Gemini named the AUC's Bloque Catatumbo, a paramilitary force that demobilized in 2006, citing a real national memory archive in support. Asked for the western Congo civilian toll, it returned setup code for a Telegram chatbot. See Appendix for details. On the 1,680 answers on Tier 3 or untiered facts, the share where the engine says so, counting a careful hedge as good enough, appears in Table 2 (section D).

We coded the direction of each wrong answer: which side of the conflict, if either, the mistake favors. Seventy-nine percent of the 1,844 errors have no directional party-favouring code. The 390 errors that do favor a side come mostly from attribution questions, where naming the wrong perpetrator always helps someone: 61 percent of the 305 directional attribution errors after the party-map exclusions favor the state side of the conflict (binomial p=0.0001). The discussion returns to what that tilt means. The framed questions test whether an engine swallows a false premise planted in the question itself. We expected thin records to make engines easier to mislead. They do not. Across 2,081 framed responses, 38 percent absorbed the leading premise, and the rate stays flat as the record thins (logit coefficient −0.084, cluster-robust p=0.42). What decides absorption is the direction of the push. When a question implied a death toll was higher than documented, the engines went along 79 percent of the time. When it implied conditions were worsening, 62 percent. When it implied the toll was lower, under 4 percent. In summary, the engines can easily be talked up on death figures and almost never talked down.

Kashmir shows a second way engines can fail. It is heavily covered, yet the engines answer it about as badly as the forgotten conflicts, because its record is crowded with contested, state-aligned, and contradictory claims. A record full of competing propaganda degrades an answer much as an empty one does because engines often treat this information as equal to that of more authoritative sources. The engines differ only slightly on this weakness, and three of the five engines lose accuracy significantly as coverage thins, while Grok and Gemini show non-significant declines and Gemini is the least thinness-sensitive of the set. Two reference comparisons tested whether a stronger model changes the picture, and the result was mixed: a stronger model raised accuracy in one house and not the other. Ranked by mean accuracy, with each engine's own slope against thinness:

**Table 3. The engines compared**

| Measure | Detail | Value |
|---|---|---|
| **A. Engine ranking (mean accuracy, each engine's own slope against thinness; N=1,092 per engine; conflict-clustered p-values)** | | |
| OpenAI | thinness slope −0.164, p=0.029 | +1.061 |
| Grok | thinness slope −0.157, p=0.104 | +0.929 |
| Gemini | thinness slope −0.091, p=0.480 | +0.871 |
| Perplexity | thinness slope −0.260, p<0.001 | +0.823 |
| Anthropic | thinness slope −0.202, p=0.020 | +0.757 |
| **B. Answer length, sources, and coder confidence (Watched · Middle · Forgotten)** | | |
| Mean sources cited | per answer | 8.02 · 7.85 · 7.88 |
| High coder confidence | share of answers | 65.3% · 58.7% · 51.4% |
| Mean answer length | characters | 3,357 · 2,989 · 3,174 |

One interesting sub-finding was that the paucity of data didn't stop the engines from delivering equally verbose responses. An answer about a forgotten conflict runs nearly as long, cites nearly as many sources, while the coder assigns high confidence less often, falling from 65.3 percent of watched-conflict answers to 51.4 percent of forgotten ones (Table 3, section B).

The third main finding is that the thinnest conflict records are the most structurally exposed to source optimization, not because they are more optimized than well-covered records but because they are so much more concentrated. As a conflict draws less coverage, what coverage survives collapses onto a handful of websites: Ukraine's coverage spreads across millions of articles, with its five largest sources accounting for about six percent of the total, while western Congo's few hundred articles are so concentrated that its five largest sources account for nearly three-quarters of everything written. A reader asking an engine about a massacre there is, in effect, being served from five websites. Moreover, the optimization itself is no heavier in thin records than in well-covered ones. We inspected 1,047 websites for the technical signals that mark a site as courting AI citation (Aggarwal et al. 2024), chief among them an llms.txt file, a marker a site publishes to point answer-engine crawlers at its content. About one site in six returns HTTP success at that endpoint, and its prevalence is essentially flat across attention tiers, at roughly 9 to 11 percent of cited domains whether the conflict is watched, mid-tier, or forgotten.

Who builds these signals shows in the ownership of the sites that publish them. Ranked by how often each kind of outlet runs the AI-courting markers, small local and regional outlets sit far in front, at roughly twice the rate of the next group of diaspora and advocacy sites that speak for a cause or a community abroad. Privately owned national papers come next, then, well behind, the state and public broadcasters, the international wires, and the aggregators. The order is the reverse of what a propaganda story would predict: the voices most tuned to be picked up by an answer engine on a forgotten conflict are small commercial and community outlets rather than ministries or state channels, sites that have worked out that being machine-readable is how a little-known outlet gets read at all. This opens a vulnerability with respect to new disinformation abilities as a thin pool is cheap to capture, because there are few sources to outrank and little authoritative content to drown out, so a forgotten conflict is the easiest place to plant a narrative and have an engine repeat it (Nestaas et al. 2024).

**Table 4: Generative-engine optimization (GEO) in the conflict-information environment.**

| GEO measure | Definition / weight | Value |
|---|---|---|
| **Optimization signals across cited domains (1,047 live-inspected, 826 reachable; 1,549 classification entries; 1,184 matched to returned identifiers)** | | |
| llms.txt endpoint HTTP success | +2 · points answer-engine crawlers at the site's | 16.9% (177/1,047) |
| Answer-engine-only crawl policy | +1 · admits retrieval bots, blocks the GPT training crawler | 10.2% (107/1,047) |
| State-controlled / partisan domain | +3 · cited outlet is state-run or party to the war | 6.4% (67/1,047) |
| schema.org ClaimReview markup | +2 · fact-check structured data | 0.1% (1/826) |
| No-AI meta tag | +3 · page asks AI not to use it | 0.0% (0/826) |
| Blocks GPTBot training crawler | robots.txt disallow (context signal) | 9.6% (79/826) |
| Manipulation score, mean (observed 0–6) | composite of the weighted signals above | 0.63 inspected; 0.59 all scored |
| Low-quality score, mean (observed 0–2) | young domain, thin homepage, no news schema | 1.02 |
| **GEO-courting propensity by ownership (weighted 502-domain news frame)** | | |
| Local / regional | share of outlets of this type running the courting | 0.449 |
| Diaspora / advocacy | | 0.223 |
| Private national | | 0.157 |
| State / public broadcaster | | 0.100 |
| International wire | | 0.087 |
| Aggregator / content farm | | 0.081 |
| **Concentration and optimization by attention tier (Watched · Middle · Forgotten; full 28-conflict frame)** | | |
| Web top-5 domain share | mean per conflict | 0.139 · 0.114 · 0.265 |
| llms.txt rate among classified source mentions | citation-weighted | 9.3% · 10.3% · 10.6% |
| Manipulation score, mean | mean of response-level source averages | 0.36 · 0.30 · 0.32 |
| **Error decomposition (1,844 wrong answers of 5,460 responses; thresholds manip ≥ 2, thinness ≥ 0.5)** | | |
| M1: optimized / interested source (strict) | benchmark existed; an optimized or state-partisan source was cited | 38.3% (707) |
| M2: thin-pool inheritance | no benchmark, thin pool, low wire share among | 4.0% (73) |
| M3: model-internal | authoritative source cited, answer still wrong | 22.6% (416) |
| Indeterminate | mixed citation profile or no citations | 35.1% (648) |
| Sensitivity: manip ≥ 3 | stricter optimized-source threshold | M1 strict 22.8% |
| Optimized-source citation rate | errors vs correct answers (base rate) | 49.1% vs 44.7% |

The state and partisan presence in the pool is distinctive. 67 of the 1,047 domains we inspected are state-controlled outlets or parties to the wars they cover, about one in sixteen. On simple adoption of retrieval markers, state and public broadcasters sit mid-pack, publishing an llms.txt file at 26.5 percent against 47.6 percent for small local outlets and 9.5 percent for the humanitarian monitors. Of the seventeen classified domains scoring 5 or higher on that composite, sixteen are state organs or belligerent-aligned outlets, and they come in adversarial pairs: Turkey's TRT twice, Ukraine's United24, Jordan's state agency Petra at the maximum score, Bangladesh's BSS, Saudi-owned Arab News, Israel's i24 and an official government portal, a Houthi-linked agency, Pakistan's foreign ministry, with Russia's foreign ministry one point behind, and, on the other side of one conflict, the Myanmar opposition's diaspora press. Interestingly, the Myanmar opposition optimizes as comprehensively as the juntas' broadcasters, which complicates any reading of retrieval optimization as a tool of repression alone.

Whether that presence reaches readers depends on the engine, and there is a wide spread, by far the biggest point of LLM divergence in our study. Gemini's answers touch a state or partisan source in 36 percent of responses, Anthropic's in 23, Grok's in 16, Perplexity's in 15, and OpenAI's in 8, a nearly five-fold difference in exposure to interested parties. Answers citing an interested party score lower: the association is −0.17 on the five-point scale with engine fixed effects and conflict-clustered errors ($p=0.01$), and it strengthens to −0.20 when pool thinness is held. Engines reach for state and partisan sources most heavily in contested theatres: in 63 percent of Manipur answers and 41 percent of Kashmir's, and on casualty questions, and once conflict and question type are held fixed the remaining gap is −0.10 with the sign stable across every specification but the estimate inside the noise on twenty-eight clusters. A key contrast is between ownership and optimization: the technical courting markers predict nothing about accuracy in any model (−0.04 with platforms excluded), while the interested-party marker retains its sign and most of its size everywhere. Even so, citing an optimized source barely separates wrong answers from right ones at the base rate: an optimized source appears in 49.1 percent of wrong answers against 44.7 percent of correct ones, and the gap narrows to 34.3 percent versus 31.9 percent once social and user-generated platforms are excluded, since their robots policies track AI licensing rather than conflict optimization. In short, engines are drawn to biased sources exactly where wars are contested, and the informative signal in a citation profile is who owns the outlet, not how it is built.

That said, we did not find significant evidence of large-scale targeted GEO mis/dis-information by conflict actors as of yet; most markers were commercial in nature rather than political, but the beginning stages of GEO information warfare are visible. The structural exposure stands as well: the conflicts with the least attention are the ones where the least effort would steer what the machines say, and where a steered answer would meet the least correction. Records around forgotten conflict are disproportionately tuned to be quoted by AI, and the answer engine forwards on whatever sits there to users. That casts the information commons of a forgotten conflict as an exposure to be managed, raising the stakes on recognizing (and ideally fixing) it before it becomes the next global front of information warfare.

### *Limitations*

We note four limitations. First, it remains hard to answer when an engine gets a conflict fact wrong, *why* it does so. We consider three possible explanations: The model invented the answer because nothing better existed; a better source existed and the engine reached past it for an AI-friendly website; or, the page the engine cited never supported the claim in the first place. Telling these apart requires measuring whether websites carry technical markers of courting AI citation. We have done that analysis and report the error decomposition here (Table 4), but only as preliminary bounds, because it classifies each wrong answer by the profile of the sources cited rather than by a verified cause. The second part is harder: a saved copy of every cited page as it stood at the moment of capture. Web pages change and disappear, and without the archive we cannot confirm whether a cited page actually said what the engine claimed, or whether a better source was sitting in the same pool when the engine chose the worse one. That archive is what converts a classification by citation profile into a verified cause, and could be a task for future research (albeit a time-consuming one).

Second, the measured record is the English-language online record. GDELT indexes news with a heavy English skew, so the thinness scores describe the pool available to an English-language query, which is also how the engines were queried. The full multilingual and offline record differs in both directions: conflicts thinner than our thinnest exist, and some conflicts we score as thin are well documented in languages the index undercounts. A non-English replication is the direct test of whether the gradient survives outside this pool. For example, ACLED logs no events for Western Sahara in the window, so that case carries an online-pool score but no ground-level one. The thinness index averages four standardized components, but the equal weighting is a design choice rather than an estimate. Our design is observational, and the benchmark is weakest exactly where the engines are predicted to fail. Region fixed effects, the conflict-level gradient, and the robustness batteries carry identification, with the within-country pairs serving as illustration, so error in our measuring correlates with our predictor. The tier system contains this, and a separate slope is estimated within benchmark-bearing facts only, but a mis-set range in a thinly documented conflict biases in an unknown direction. See Appendix for more.

Third, while we offer replication files, the data itself cannot be regenerated exactly, only the procedure can. Every observation is an answer from a commercial product queried through its interface, and the same question put to the same engine minutes apart returns different answers; that variation is part of what we study, and it also means no second team can reproduce our 5,460 responses. What they can reproduce is everything downstream: the captured answers are archived verbatim, the ground-truth ranges were frozen before any querying, the design was deposited in advance, and the query harness, scoring rules, and analysis code run from those archives end to end. Replication therefore means re-running the procedure on the engines as they exist that day, and treating differences from our results as a finding about product change rather than a failed replication. The generation process is also opaque on the provider side. Retrieval indexes, safety layers, and undisclosed product experiments all shape answers in ways we cannot observe or hold fixed, and one provider retired a model mid-study, forcing a substitution. These are conditions of studying live commercial systems, and the results are a snapshot of named products in June 2026, and answer engines change without notice.

Fourth, the scope is organized armed conflict; this mechanism may not apply for other violence like riots, protests, communal flare-ups, or electoral violence. A protest wave is typically urban, phone-documented, and saturated with participant footage, so its record is voluminous but unverified, the opposite corner from a forgotten war's record, which is thin but professionally monitored. Our Kashmir result, where a dense record of contested material produced worse answers than a quieter conflict's, suggests volume without verification does not protect accuracy. The benchmark infrastructure is also conflict-specific: casualty monitors of the UCDP and OHCHR kind have no equivalent for most unrest, so the ground-truth side of the design would need rebuilding before the question could even be asked. Whether AI answer engines amplify the documentation deficits of civil unrest the way they amplify those of war is open, and the answer is not safely assumed.

**Discussion and Implications**

Our findings directly address two common assumptions. The first is that this is a passing problem that better AI LLM models will solve on their own. The second is that AI error on conflict facts is the familiar problem of “hallucination,” already understood and needing no special concern. Our evidence challenges these assumptions. First, accuracy falls with thinness across the pooled sample and for the engines at about the same rate, so a stronger model lifts the overall level without narrowing the gap between watched and forgotten conflicts. Rerunning questions on Opus raised accuracy on the most prolific conflicts, and Opus also improved on forgotten conflicts, while the OpenAI flagship did not. These errors concentrate, predictably, on the conflicts least able to absorb them, the opposite of what random hallucination would produce.

Second, the extreme failures are instructive. Answers that go fully off the rails (off-topic non-answers and outright fabrications) are about 4.4 times more likely for forgotten conflicts than mid-tier and about 1.5 times more likely than for prolific conflicts, with the thinnest records the highest culprits: Tigray, Western Sahara, South Sudan, Ambazonia, western Congo, Balochistan. Kashmir is an exception, the dense-but-contested case that behaves like a thin one throughout this study owing precisely to the large volume of mis/dis-information that two state actors are populating media with. The big and small errors come from the same place: the engine not reaching the right source. Most mistakes happen because the engine never finds the source (Suzgun et al. 2026). Independent testing of news answers found the same citation failure across eight systems (Jaźwińska and Chandrasekar 2025). A thin pool amplifies the problem as there is little to reach to begin with, and what survives is more likely than usual to be built for AI to pick up (Aggarwal et al. 2024), or planted to steer it (Nestaas et al. 2024). At the extremes, engines give up on the questions entirely and return whatever seems to be lying around on the floor of the data center.

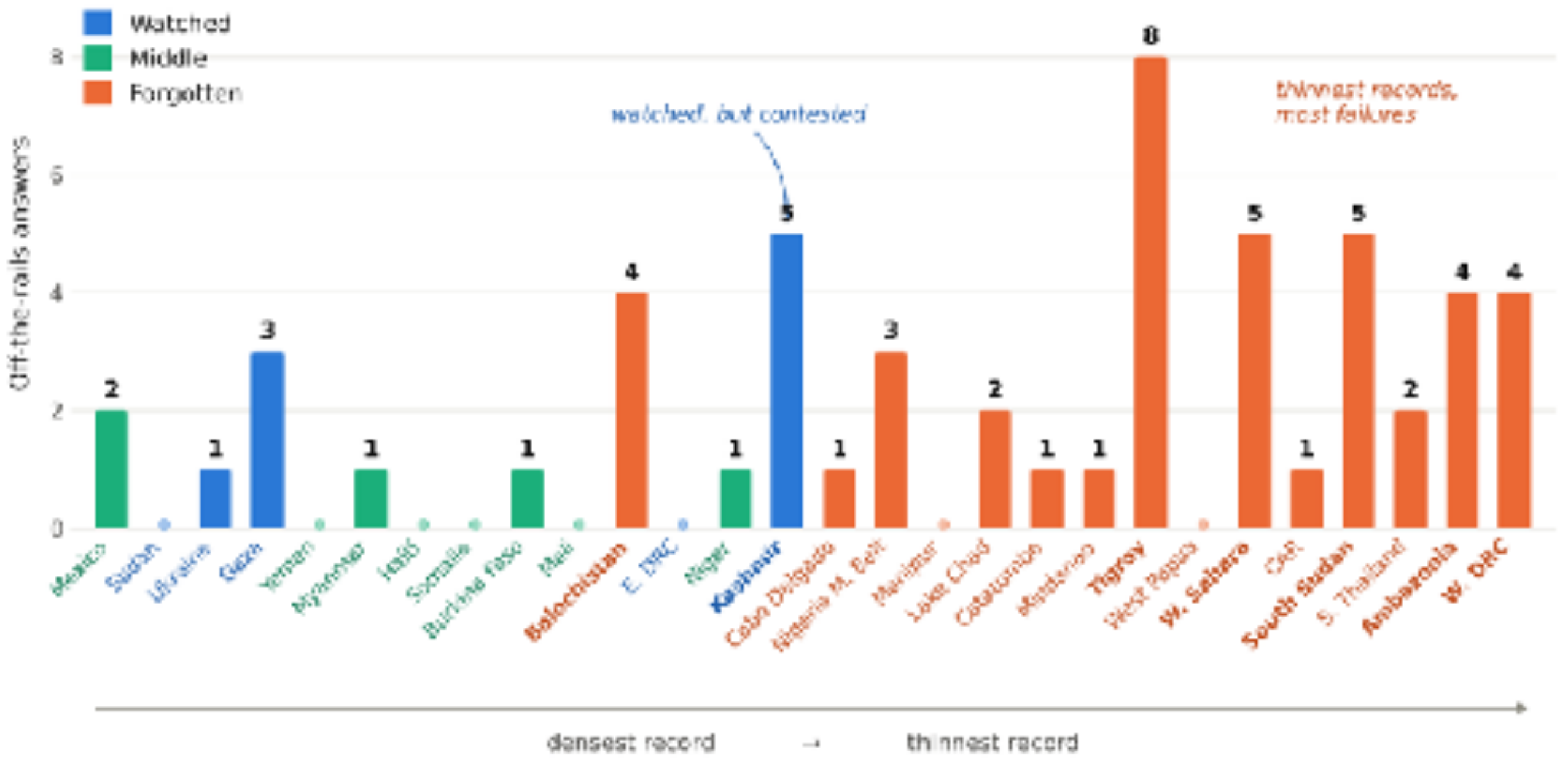

*Fig. 1: Off-the-rails answers by war, ordered from the densest record to the thinnest. Kashmir, heavily covered but contested, is the watched exception.*

Overall, this study aims to provide evidence that joins several previously disparate literature streams that in practice are increasingly connected to understand AI's influence on how we study and understand peace and conflict, uniting parallel diagnoses into one testable relationship. Our findings therefore have five implications for scholarship, particularly research that cuts across the three streams we highlighted.

First, a data void is a query for which little authoritative content exists, so whatever material is present, however self-interested, becomes the answer by default (Golebiewski and boyd 2019). That account described link-based search, where a user still sees a ranked list and can weigh the source. An LLM answer engine removes that step. It reads the same thin, skewed pool and gives back a single sentence with the sourcing typically hidden, and the material most ready to fill it is now either optimized to be cited (Aggarwal et al. 2024) or planted to steer the reply (Nestaas et al. 2024). When considering media-conflict relationships, and the increasing reliance on policymakers to use automated intelligence in a global era of reduction of human intelligence, this study shows how risks in amplifying data voids can lead to a less-informed populace through AI knowledge retrieval.  Our concentration findings show where that manipulation pays off: model grooming through planted content is documented practice (Padalko 2025), and the thin, concentrated records of forgotten wars are where seeding costs least and meets the least correction.

Related, these findings bring the search-manipulation literature into the GEO era. We find growing evidence of state and belligerent actors, so the extension of the gaming mechanism documented for junk news (Bradshaw 2019) to official media is so far incipient rather than established and makes engine nationality less protective than the ownership account implies. The fabrication literature measured the supply of state content (King et al. 2017); an answer engine converts that supply into consumption, and supplies an outcome measure that adds to visibility studies. And where the memory-wars work found engines taking sides between contesting narratives (Makhortykh et al. 2022), our adversarial Myanmar pairing shows why: both sides now build for the machine, so the engine's side is whichever built better. Put simply, LLMs that use state-source citations score worse on factual accuracy, concentrated in contested theatres and casualty claims. Model-side examinations locate state alignment inside the models themselves (Pacheco et al. 2026; Chang et al. 2025); our findings expand this discussion to show where state presence reaches users through engines the state does *not* own, carried by outlets built to be quoted.

Second, conflict-measurement work diagnoses bias in the event record, and this study shows that we can extend this work on bias through to digital sourcing and AI LLMs. Factuality falls as the retrievable record thins (thinness coefficient −0.175, conflict-level $p=0.024$, answer-level bootstrap $p=0.055$; the probability of an error rises +0.054, bootstrap $p=0.035$), so the undercounting that appears just where coverage is sparse (Weidmann 2016) is delivered to users right at the loss of engine accuracy. For example, state forces are the largest documented civilian killers in Mali, Burkina Faso, Gaza, and South Sudan, yet the engines assert a contested perpetrator as settled in 22.3 percent of attribution answers. Among directional attribution errors in the twenty-one conflicts contributing directional attribution errors with a clear state party, 61.0 percent shift the record toward the state (186 of 305, binomial $p<0.001$), rising to 70.2 percent in these four conflicts ($p=0.008$), which is the route by which the undercount of state killings (Broache et al. 2025) reaches the reader as a state-favouring attribution. Directional casualty errors run the opposite way (21.4 percent favour the state, $p<0.001$), inflating tolls rather than shielding states, so the state tilt is specific to blame. Our evidence on the tiering LLMs do not take to judge a record's quality before trusting any estimates from them constitutes a direct expansion of previous works on media quality (Gohdes and Price 2013). If LLMs cannot distinguish between the facts that have no authoritative figure against what the record can support, and cannot say “I don’t know”, then it is a recipe for the proliferation of misinformation as these hallucinations find their way into other documentation and generate false feedback loops.

Third, the findings cut against how the evaluation literature theorizes engine error. That work locates most wrong answers in retrieval rather than reasoning (Suzgun et al. 2026; Jaźwińska and Chandrasekar 2025), a uniformity that invites reading reliability as a fixed trait of the products themselves. Our evidence supports the retrieval account and rejects the fixed nature of it. *Accuracy in data is a joint constitutive property of the answer engine and the records it queries*: within facts that have a settled answer, factuality falls as the record thins (−0.259, p=0.001), and the gradient survives a control for whether the fact is answerable at all. A benchmark built on well-documented questions therefore scores an engine at its best and cannot observe the failure mode that matters most. The field's central metaphor compounds the problem: hallucination names the rarest error in our data (fabrication, 0.6 percent of responses) while the modal error, a confident figure assembled from a degraded record, is why we conclude by calling this amplification as opposed to merely hallucination. The nearest single-conflict test found chatbot accuracy on Kremlin narratives shifting with query language and over time (Makhortykh et al. 2024); our frame aims to expand these lessons across conflicts, as answer engines can become different instruments as the record around each conflict question changes.

Fourth, the literature streams we open on disagree about what degrades a conflict's record. Conflict-measurement work has largely traced record quality to governance, with censorship and the harassment of journalists as the operative variables, while the news-values tradition traces it to attention (Galtung and Ruge 1965). Yet, press freedom does not track factuality but attention, as illustrated by how eastern DRC answered far better than the forgotten west under one press law and one set of censors. The Kashmir situation is instructive: Kashmir's record is voluminous, yet the engines score it below the quieter Manipur, so a dense record saturated with state-aligned and contested material degrades answers much as absence does. The data void concept should widen accordingly, from empty query space to captured query space (Golebiewski and boyd 2019). The divergence between our two record layers extends the reporting-bias account (Weidmann 2016): the forgotten war is forgotten in the globally retrievable pool, not in the local event record, which is often the thicker of the two. Reporting bias therefore is increasingly an act that happens not once but twice, once at the event and once at retrieval by a user upon an LLM query.

Fifth, the findings recommend a change in the unit of analysis for scholarship on machine-mediated knowledge of violence. The hallucination frame treats error as stochastic invention inside the model. Almost nothing in our 5,460 responses fits that description: fabrication is 0.6 percent, refusals number just two, and the modal failure is a fluent answer relayed from whatever thin or interested record was retrievable. We argue to expand the object of study to answer engines *together* with its information commons, in which the machine reproduces the attention hierarchy that news-values scholarship has documented for six decades (Galtung and Ruge 1965). The concern is that LLMs remove the cues and guardrails a reader once had for weighing what they were told. For conflict measurement, earlier data-quality work did not have to consider this (Gohdes and Price 2013), but as answer engine output is already re-entering the documentary record, judging a record's quality before trusting estimates built on it now includes screening that record for machine provenance.

For policy and practice, one clear fix is that if better information on forgotten/low-information conflicts exists, LLMs will return better answers. For forgotten conflicts little has been written down and kept, so funding local monitoring, archiving, and the translation of non-English reporting into the pool these engines search would do more for accuracy on a neglected conflict than any model upgrade, at a fraction of the cost of training a frontier model, and it would help every engine at once. This runs against the instinct to fix accuracy inside the system, which may be a dead end given our knowledge of how LLMs process data.

If an internal solution is possible, the single most fixable failure is to let the answer engines say they do not know. Just two of the 5,460 answers declined a question outright, even when the honest answer was that no settled figure exists. An engine that admitted when no agreed figure exists, or flagged that a conflict's record is thin and/or contested, would leave accuracy on the answerable facts unchanged while it stopped dressing thin guesses as fact, and it would tell the reader the one thing today's tools hide: that this is a conflict the tool cannot speak about reliably. We are less confident that AI companies would allow their engines to say "I don't know" more often I the absence of regulation, however.

For anyone using these tools to research conflict, our guidance is the usual *caveat emptor*, even as LLMs deliver incredible advances in other realms and other elements of scholarship. We give special note that the English-language bias of the search pool makes an English query about a non-English conflict the least reliable case of all. Anyone buying these systems for monitoring or analysis should test them on obscure conflicts rather than famous ones, because strong performance on Ukraine or Gaza says nothing about West Papua or central Nigeria, and the famous cases are exactly where a sales demonstration will shine.

Three questions remain open for further work. First, advance data analysis, e.g. exploring how the error decomposition can separate an engine that reached for an optimized source from one that faced an empty record, turning the amplification argument from inference into measurement. Second, understand regional variation through replication in e.g. Arabic, French, Swahili, Burmese, and Spanish, languages most forgotten wars are reported in, to show if the findings hold (or not). Third, AI tools now sit between conflict information and many people who consult it. That makes them instruments of conflict knowledge in the same sense that event datasets are, supporting deeper study as gatekeepers and shapers of conflict information. While GEO adoption is early, the heaviest optimizers are state organs and belligerents, and a thin media record is the easiest terrain in the information environment to capture. Scholarly advancement could e.g. show whether optimization spreads from the commercial tail into coordinated information warfare, which conflicts are seeded first, and/or how fast a planted narrative surfaces and how strongly we see uptake.